\documentclass[letterpaper, 10 pt, conference]{ieeeconf}  

\IEEEoverridecommandlockouts                              
\usepackage{graphicx} 
\usepackage{times} 
\usepackage{amsmath} 
\usepackage{amssymb}  
\usepackage{verbatim} 
\usepackage{cite}
\usepackage{subfigure}
\usepackage[hyphens]{url}
\usepackage[ruled,vlined]{algorithm2e}
\usepackage{float}
\usepackage{tabulary}
\usepackage{tabularx}
\usepackage{lipsum,amsmath,multicol}
\usepackage{tensor}
\usepackage{multirow}
\usepackage{booktabs}
\usepackage[disable]{todonotes}
\usepackage[nolist]{acronym}
\usepackage{tikz}
    
\usepackage{lipsum}
\usepackage{siunitx}
\def\fps@figure{htp}
\def\fps@table{htp}

\newcommand{\bi}{\begin{itemize}}
\newcommand{\ei}{\end{itemize}}

\newcommand{\bfig}{\begin{figure}}
\newcommand{\efig}{\end{figure}}

\newcommand{\benum}{\begin{enumerate}}
\newcommand{\eenum}{\end{enumerate}}

\newcommand{\be}{\begin{equation}}
\newcommand{\ee}{\end{equation}}

\newcommand{\ba}{\begin{eqnarray}}
\newcommand{\ea}{\end{eqnarray}}

\usepackage{color}

\definecolor{CommentRed}{rgb}{0.7,0,0}
\definecolor{CommentBlue}{rgb}{0,0,0.7}
\definecolor{CommentDG}{rgb}{0,0.6,0}

\newcommand{\renaud}[1]{\todo[inline,color=green!40]{\textbf{Renaud:} #1}}

\makeatletter
\newenvironment{myalign*}{%
  \setlength{\mathindent}{0pt}%
  \setlength{\abovedisplayskip}{-\baselineskip}%
  \setlength{\abovedisplayshortskip}{\abovedisplayskip}%
  \start@align\@ne\st@rredtrue\m@ne
}%
{\endalign}
\makeatother

\title{\LARGE \bf
Design of an Autonomous Racecar: Perception, State Estimation and System Integration
}

\author{$\text{Miguel I. Valls}^{*},\text{Hubertus F.C. Hendrikx}^{*} , \text{Victor J.F. Reijgwart}^{*},  \text{Fabio V. Meier}^{*}$ \\ \text{Inkyu Sa}, \text{Renaud Dub\'e}, \text{Abel Gawel}, \text{Mathias B\"urki}, and \text{Roland Siegwart}\thanks{${}^{*}$ The authors contributed equally to this work.}
\thanks{Authors are with the Autonomous Systems Lab, ETH Zurich, Zurich, Switzerland. \texttt{dmiguel@ethz.ch}}%
}

\begin{document}

\maketitle

\thispagestyle{empty}
\pagestyle{empty}


\begin{abstract}
This paper introduces \emph{fl\"uela driverless}: the first autonomous racecar to win a Formula Student Driverless competition. In this competition, among other challenges, an autonomous racecar is tasked to complete 10 laps of a previously unknown racetrack as fast as possible and using only onboard sensing and computing. The key components of \emph{fl\"uela}'s design are its modular redundant sub--systems that allow robust performance despite challenging perceptual conditions or partial system failures. The paper presents the integration of key components of our autonomous racecar, i.e., system design, EKF--based state estimation, LiDAR--based perception, and particle filter-based SLAM. We perform an extensive experimental evaluation on real--world data, demonstrating the system's effectiveness by outperforming the next--best ranking team by almost half the time required to finish a lap. The autonomous racecar reaches lateral and longitudinal accelerations comparable to those achieved by experienced human drivers.
\end{abstract}

\section{INTRODUCTION}
\label{sec:intro}

On August 13th 2017, \emph{fl\"uela driverless} became the first car to ever win the Formula Student Driverless (FSD) competition. 
The competition requires the car to race fully autonomously and consists of 4 dynamic and 4 static disciplines \cite{fsg_rules}. The dynamic disciplines test the system's reliability under general race conditions and at high lateral and longitudinal speeds. The static disciplines evaluate the system's design under aspects of software, hardware, costs, and business. While \emph{fl\"uela driverless} performed well in all categories, we this paper focuses on software and hardware designs.

The hardware platform for the project is \emph{fl\"uela}, an electric 4WD car with a full aerodynamic package, high wheel torque, and a lightweight design developed by AMZ\footnote{\url{http://www.amzracing.ch/}} for Formula Student Electric 2015. 
The sensor outfit for autonomous operation and the software system are developed from scratch.

In our autonomous design, system reliability under high performance operation is chosen as the main design goals, since the FSD regulations allow no human intervention.

\begin{figure}
\begin{center}
\includegraphics[width=\columnwidth]{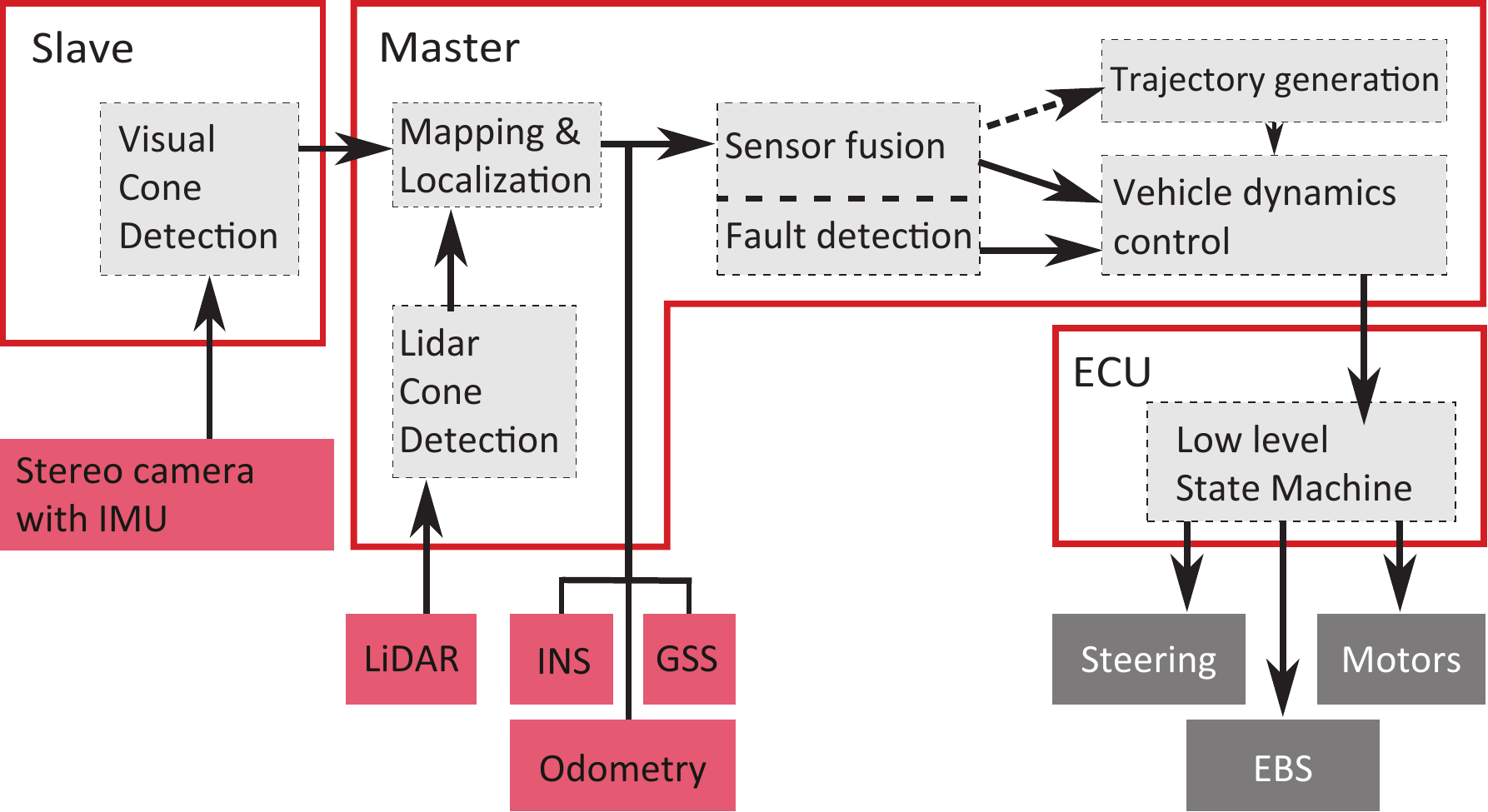}
\end{center}
    \vspace{-5mm}
    \caption{Overview of the autonomous system's architecture.}
    \label{fig:SystemArchitecture}
    \vspace{-6mm}
\end{figure}

This paper presents the state estimation, LiDAR SLAM, and localization systems that were integrated in \emph{fl\"uela}.
The autonomous system perceives its surroundings using a 3D LiDAR and a self-developed visual-inertial stereo camera system. Furthermore, a velocity sensor and an Inertial Navigation System (INS) combining an IMU and a GPS were added for state estimation. All the information is processed online by two computing units running Robot Operating System (ROS) and a real-time capable \ac{ECU}. Figure \ref{fig:SystemArchitecture} shows an overview of the hardware-software setup.

In order to reach \emph{fl\"uela}'s full potential when racing autonomously, the track must be known at least $2s$ ahead. At high speeds, this requires a perception horizon that exceeds the sensors' range. The car must thus drive carefully to discover and map the track. This mode will later be referred to as \textit{SLAM Mode}. Once the map is known, the car can drive in \textit{Localization Mode} which can exploit the advantage of planning on the previously mapped race-track.\\
The contributions of this paper are:
\begin{itemize}
\item A complete pipeline from perception to state estimation with on-board sensors and computation only, capable of driving an autonomous racecar close to a human driver's performance.
\item Extensive experimental evaluation and demonstration in real-world racing scenarios.
\end{itemize}

\begin{figure}
\begin{center}
\includegraphics[width=\columnwidth]{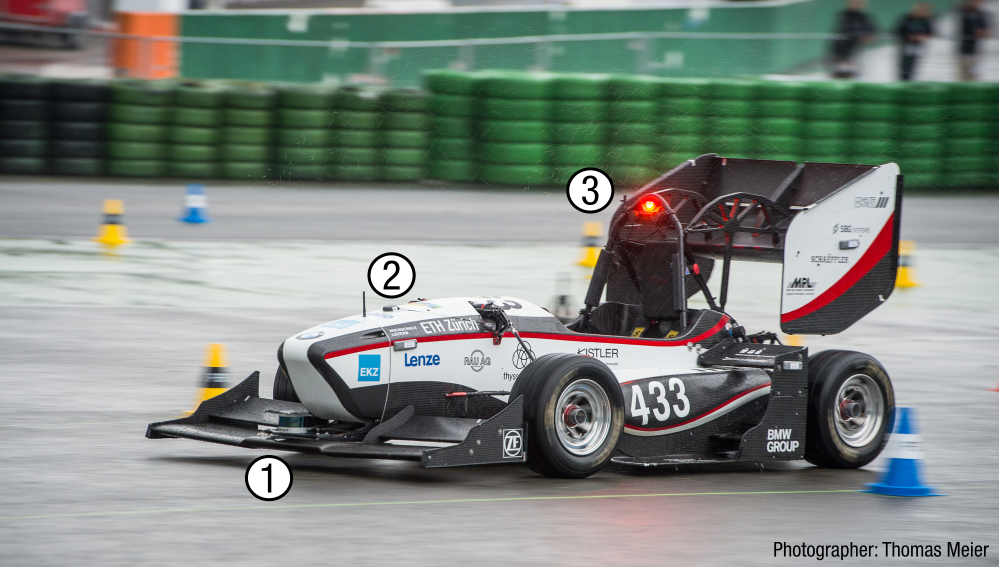}
\end{center}
    \caption{\emph{fl\"uela driverless}, the winning electric, autonomous 4WD racecar at the Formula Student Germany 2017. The LiDAR, the GPS and the visual-inertial system are respectively marked by tags 1 to 3. \renaud{Do you have a strong opinion about having the f of fluela as lowercase? (miv) yes we do, it's an AMZ style decision to do this for all cars}}
    \label{fig:M100}
    \vspace{-5mm}
\end{figure}

The remainder of this paper is structured as follows. Section \ref{sec:background} introduces state-of-the-art work on autonomous racing, Section \ref{sec:methodologies} describes the theoretical development for this project and Section \ref{sec:implementation} the implementation details. We present our experimental results in Section \ref{sec:results}, and conclude in Section \ref{sec:conclusion}.

\vspace{-5pt}

\section{Related Work/Background}\label{sec:background}

Autonomous racing is an emerging field within autonomous driving. In the last years, a few self-racing vehicles have been developed, both in academic and in the industrial research. The first known autonomous vehicle competition was the DARPA Grand Challenge, \cite{DGC} which motivated the development of several autonomous cars in a two year period. These cars had to compete in a desert environment and drive through a way-point corridor given shortly before the race. In this sense, it is similar to FSD since a short period for mapping is allowed just before the race. They however differ in that the FSD track is asphalt, the vehicles are designed for racing and reached over $90 km/h$ and $10 m/s^2$ accelerations.

Other autonomous racecars were developed afterwards \cite{audiTT}, but their main goal was vehicle dynamic control and not SLAM and state estimation. In addition, several scaled racecars were developed \cite{orca} but they focus on control and have an external localization system. Others were developed with on-board sensors only \cite{barc} but the main focus also lied on control. 

Finally, a look to the industry should not be forgotten, where an Audi RS7 and a Nio EP9 broke the speed record for autonomous cars in 2014 and 2017 respectively. Devbot from roborace also featured the first wheel to wheel autonomous race (2017).

\section{fl\"uela driverless}
\label{sec:methodologies}
In this section an insight is given in the full setup of the state estimation system, including the LiDAR/camera based mapping \& localization system. First a high level system overview is presented.
\subsection{System architecture}
\label{sec:system-architecture}
The system architecture has been designed for reliability and performance. Reliability was given largest priority as the competition only grants one run, regardless of adverse weather conditions or software glitches.

The car is fitted with an Inertial Navigation System, an optical Ground Speed Sensor (GSS), a LiDAR and a self-developed visual-inertial stereo camera system. Furthermore, individual wheel speeds are determined by reading out each wheel encoder.
Consumer-grade GPS is used (no differential GPS or RTK) as an absolute position sensor.
Cones that mark the race-track are detected by both LiDAR and camera to create redundancy in the perception pipelines.

The chosen computing system consists of a high performance slave and an industrial master computer. The slave computer is dedicated to vision-based perception and the master computer runs all other software packages. Since vision-based perception is redundant with LiDAR, this solution ensures high reliability without limiting performance. The last important factor for reliable operation is the self-developed computing housing, presented in Sec.~\ref{sec:computing-housing}.

The designed software system runs on Ubuntu 14 LTS within the ROS Indigo framework. The distributed nature of ROS simplifies the integration of the slave computer. Chrony is used to synchronize the clocks of both computers over Ethernet.

Finally, a real-time capable ECU runs the low level controllers and low level state machine of the car. The torque vectoring and traction controllers developed for the original car are used to distribute individual torques to all 4 wheels at $200\si{\hertz}$. Their target is the desired throttle calculated on the master computer. The car relies on regenerative braking encoded as a negative throttle input during normal operation and the mechanical brakes are only used for emergency stops. Lastly, the ECU forwards the desired steering angle from the master computer to the internal controller of the steering actuator after a simple integrity check.
\subsection{Pose and Velocity Estimation}
\label{sec:state-estimation}
State estimation is an essential part of any mobile robotic application as it enables the robust operation of other system components.
Several sensors are fused to estimate the pose and velocity of the ground vehicle.
To take advantage of redundancy in state estimation, the contribution of each sensor input to the overall estimated state has to be quantified in function of 
the sensor's accuracy and previous state knowledge.
The Extended Kalman Filter (EKF) is the state-of-the-art estimator for fast, mildly non-linear systems. For systems with white zero-mean additive gaussian noise corrupting the sensors and the process model, it is a good approximation of the optimal solution.

\subsubsection{Process model}
The process model used is driven by the accelerometer as proposed in \cite{msf_paper}. The vehicle body frame is chosen to coincide with the IMU frame. A constant velocity model is used with the accelerometer as a pseudo input to the system. Due to the application constraints, it is known that the vehicle will remain on the ground and not be substantially tilted. This assumption simplifies the state to a 2D state with only 6 elements. 
The state vector is defined as:
\begin{align}\label{eq:state}
\begin{split}
\mathbf{x} &= [\mathbf{p}^T, \,\theta, \, \mathbf{v}^T, \, r]^{T} \\
\mathbf{p} &= [x, \, y]^T\\
\mathbf{v} &= [v_x, \, v_y]^T
\end{split}
\end{align}
where $\mathbf{p}$ and $\theta$ are respectively the position and heading of the car (IMU) expressed in world reference frame.
$\mathbf{v}$ and $r$ are respectively the linear and angular velocities of the car expressed in body reference frame.
The process model is defined as:
\begin{align}
\begin{split}
\dot{\mathbf{p}} &= \mathbf{R}(\theta)^{T} \mathbf{v} \\
\dot{\theta} &= r \\
\dot{\mathbf{v}} &= \mathbf{a} + \left[ v_y r, \, -v_x r \right]^T+ \mathbf{n_v}\\
\dot{r} &= n_r
\end{split}
\end{align}
where  $\mathbf{a}$ is the linear acceleration measured by the IMU, $\mathbf{R}(\theta)$ is the 2D rotation matrix between the vehicle body frame and the world reference frame,  $\mathbf{n}_v$ and $n_r$ are i.i.d white noise distributed as $\mathbf{n}_v \sim \mathcal{N}(0,\,\mathbf{\Sigma}_{v}) $, $n_r \sim \mathcal{N}(0,\,\Sigma_{r}) $.
\subsubsection{Sensor model}\label{sec:sensor_model} 
The vehicle is equipped with multiple sensors (see Sec. \ref{sec:intro}) which can be decomposed on the quantities being measured: position ($\mathbf{z}_{p}$), heading ($z_{\theta}$), velocity ($\mathbf{z}_{v}$), and yaw rate (${z}_{r}$). \renaud{I am not sure about this. It basically says that a position sensor is $z_p$.  answ: I don't understand} 
\renaud{I am surprised that the reviewers did not comment about this. Variable $z_s$ usually refer to measurements.. not the sensors. I would propose the following:  The vehicle is equipped with multiple sensors (see Sec. \ref{sec:intro}) which are differentiated based on the quantity being measured: position ($\mathbf{z}_{p}$), heading ($z_{\theta}$), velocity ($\mathbf{z}_{v}$), and yaw rate (${z}_{r}$). This is not perfect but at least it is not wrong. (miv) is it better now?}
For instance, the GPS can be seen as a position sensor and the localization described in (Sec. \ref{sec:mapping-and-localization}) as a position and a heading sensor. For this to hold, it is assumed that the noises of the decomposed measurements are uncorrelated. 
The measurements are introduced as 
\begin{align}\label{eq:sensor_model}
\begin{split}
\mathbf{z}_{p} &=  h_{p}(\mathbf{x}) = \mathbf{p} + \mathbf{R}(\theta)^T \mathbf{p}_{s} + \mathbf{n}_{z_{p}}\\
z_{\theta} &= h_{\theta}(\mathbf{x}) = \theta + \theta_{s} + n_{z_{\theta}}\\
\mathbf{z}_{v} &= h_{v}(\mathbf{x}) =   \mathbf{R}(\theta_{s}) (\mathbf{v} + [-r\, \mathbf{p}_{s, y}, \, r\, \mathbf{p}_{s, x}]^T) + \mathbf{n}_{z_{v}}\\
z_r &= h_{r}(\mathbf{x}) = r + n_{z_{r}}
\end{split}
\end{align}
where $h_{\{\cdot\}}(\mathbf{x})$ are the different measurement models, $\mathbf{p}_{s}$ is the position of the sensor in body frame and $\theta_{s}$ is the sensor heading in body frame. $n_{\{\cdot\}}$ are gaussian i.i.d. noises that corrupt the sensor measurements.
\subsubsection{Observability analysis}
In order to determine for which states the system is observable, the observability matrix of the non-linear system must be analyzed. It can be constructed using the Lie derivatives of the sensor model presented in \ref{sec:sensor_model}. They are defined recursively as
\begin{equation}\label{eq:Liederivative}
L^{l+1}_{f} h (\mathbf{x}) = \frac{\partial L^{l}_{f} h}{\partial \mathbf{x}} f(\mathbf{x}, \mathbf{u})
\end{equation}
with $L^{0}_{f} h (\mathbf{x}) = h(\mathbf{x})$. \\  
The Observability matrix is defined as
\begin{equation}\label{eq:Observavilitymatrix}
\mathcal{O}(\mathbf{x}, \mathbf{u})=\left[ \frac{\partial L^{0}_{f} h (\mathbf{x}) }{\partial \mathbf{x}} , \, \frac{\partial L^{1}_{f} h (\mathbf{x})}{\partial \mathbf{x}}, \, ...   \right]^T
\end{equation}
By performing a rank-test on $\mathcal{O}$, it can be determined whether the system is weakly locally observable (in case of full column rank,  \cite{nonlinear_observability}) or not observable.
This analysis yields three scenarios:
\begin{enumerate}
\item The state is observable if there is at least one position and one heading sensor.
\item The state is not observable if there is no position sensor.  
\item The state is observable except at stand-still if there is a position sensor but no heading sensor.
\end{enumerate}
In the current setup, there always is a position sensor (GPS) but no heading sensor until the map is known and localization output is fed to state estimation, which means scenario 3) in \textit{SLAM Mode} and 1) \textit{Localization Mode}. 
To overcome the fact that the heading cannot be estimated at stand-still if the map is not known, a Frozen Pose Update (FPU) is implemented. It differs from the Zero-velocity update (ZUPT) since it assumes a constant pose instead of zero velocities. As long as zero-motion is detected, a virtual measurement is added that simulates a position and heading sensor with the value of the frozen pose. This prevents the system from drifting even if the process model is biased or there is noise in the velocity sensors. In this application, the whole pose is frozen instead of only the heading as it is more important for the pose estimate to not drift at stand still than to drift towards the actual position.

\subsubsection{Delay compensation}
The EKF approach can only take into account measurements from the current state which is a problem with delayed measurements.
A trivial solution is to keep a buffer of previous state distributions and measurements, and at every iteration the state is propagated forward and corrected with all the newer measurements up to the current time.\\
Although this approach is simple, consistently delayed measurements considerably increase the computational time of the filter. For other methods addressing discrete Kalman filters with delays, see \cite{delayed_DKF}.
\begin{figure}
\begin{center}
\includegraphics[width=1\columnwidth, trim=.8cm .5cm .8cm .5cm, clip]{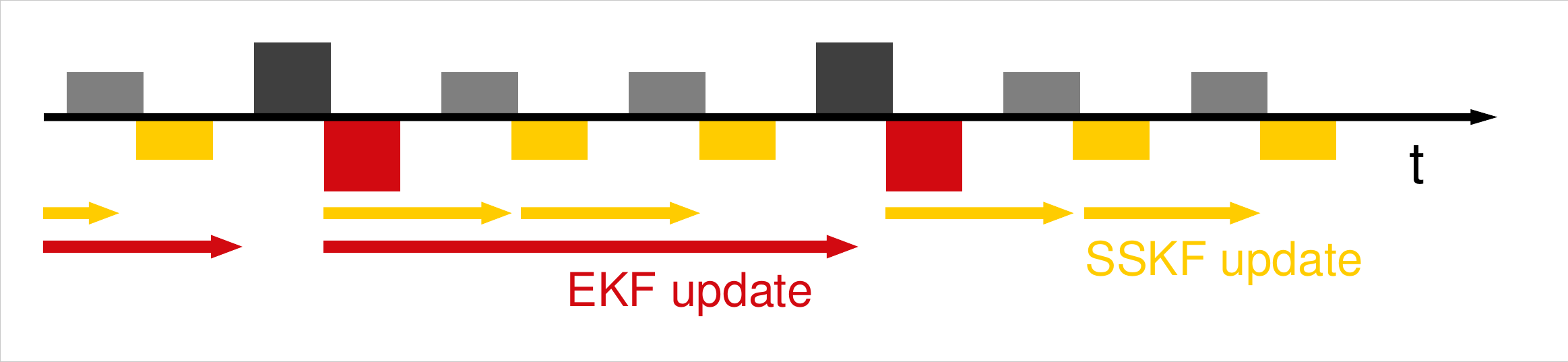}
\end{center}
    \vspace{-5mm}
    \caption{Approximate delay compensation. The EKF accurately estimates the car state (red) up to the most recent low frequency measurement (black). Fast, non-delayed measurements (grey) are incorporated into a high frequency, temporary estimate (yellow) using the SSKF. This method locally approximates the model as an LTI system with stationary noise distributions.}
    \label{fig:delay_compensation}
    \vspace{-6mm}
\end{figure}
We propose an approximate approach (see Fig. \ref{fig:delay_compensation}) where the EKF is calculated up to the most delayed measurement. A steady-state approximation of the EKF (SSKF) is then executed, taking into account all measurements newer than the most delayed one 
to keep a high-rate updated estimate for the control system. The SSKF is a simplified version of the EKF, where the covariance is assumed to be constant (or slowly varying) for the interval from the most delayed measurement to the current time. The measurement model is assumed to be close to linear and the measurement noise and process noise are assumed to be stationary for this interval. This leads to a constant Kalman gain, avoiding the matrix inversion step. There is also no need to calculate the covariance in this interval. 
This approach provides a trade-off for systems with delayed measurements that balances the accuracy of the EKF and runtime of SSKF. 

\subsubsection{Outlier rejection and self-diagnosis}
\label{sssec:outlier_rejection}
Sensor faults are a major factor undermining the robustness of state estimation systems.
We therefore use a probabilistic outlier detection method that works with any sensor. The idea was first presented by Brumback and Srinath~\cite{outlier_detection_original_paper} and later used by Hausman et al.~\cite{outlier_detection_paper}. 
This approach makes use of the innovation covariance calculated in the EKF. This allows one to assess the likelihood of a measurement belonging to innovation distribution. 
This approach intrinsically accounts for the uncertainty of the state and the sensor noise model:
\begin{align}
\mathbf{r} &= \mathbf{z} - h(\hat{\mathbf{x}}) \\
\mathbf{S} &= \mathbf{H}\mathbf{P}\mathbf{H}^T+\mathbf{R}
\end{align}
where $\mathbf{r}$ and $\mathbf{S}$  are the residual (or innovation) and its covariance. $\mathbf{z}$ and $\mathbf{R}$ are the measurement and its covariance. $\hat{\mathbf{x}}$ and $\mathbf{P}$ are the estimated state and its covariance. $h(\cdot)$ is the measurement model and $\mathbf{H}$ is its linearization around $\hat{\mathbf{x}}$. 

If the Chi-Squared ($\chi^2$) test fails, the measurement is considered an outlier.
It fails when Eq. \ref{eq:chi2test} does not hold:
\begin{equation}\label{eq:chi2test}
\mathbf{r}^T\mathbf{S}^{-1}\mathbf{r} < \chi^2(\chi_i)
\end{equation}
 where $\chi^2(\cdot)$ is the Chi-Squared distribution of as many degrees of freedom as the size of $\mathbf{r}$ and $\chi_{i} \in (0,1)$ is the threshold to reject an outlier of the $i^{th}$ sensor in the $\chi^2$ test.

In this paper, the outlier detection is extended to \textit{health} estimation and diagnosis based on the same idea as the outlier detector (the normalized sum of squares of the residuals). This normalized sum is scaled to reach 1 when it is considered an outlier and saturated to 1.
\begin{align}
D_i &= 1 - max \left\{ \frac{ r^{T}_i S^{-1}_{i}r_i }{\chi^2(\chi_i)} \, , \, 1 \right\}\\
D_T &=\frac{\sum_{i=1}^{p} w_i \,  D_i}{\sum_{i=1}^{p} w_i}
\end{align}
where $D_i \in [0,1]$, $r_i$ and $S_{i}$ are the last diagnosis, last residual and last innovation matrix of the $i^{th}$ sensor respectively. $w_i$ is the weight of $i^{th}$ sensor in the weighted sum that determines the overall diagnosis of the system ($D_T$). $p$ is the number of sensors.
$D_T \in [0,1]$ where 0 means that every sensor is an outlier and 1 means that every sensor is perfectly predicted. The weights $w_i$ are introduced to represent the impact of every sensor on the overall health diagnosis of the system. 

\subsection{LiDAR Cone Detection}
\begin{figure}
\begin{center}
\includegraphics[width=\columnwidth]{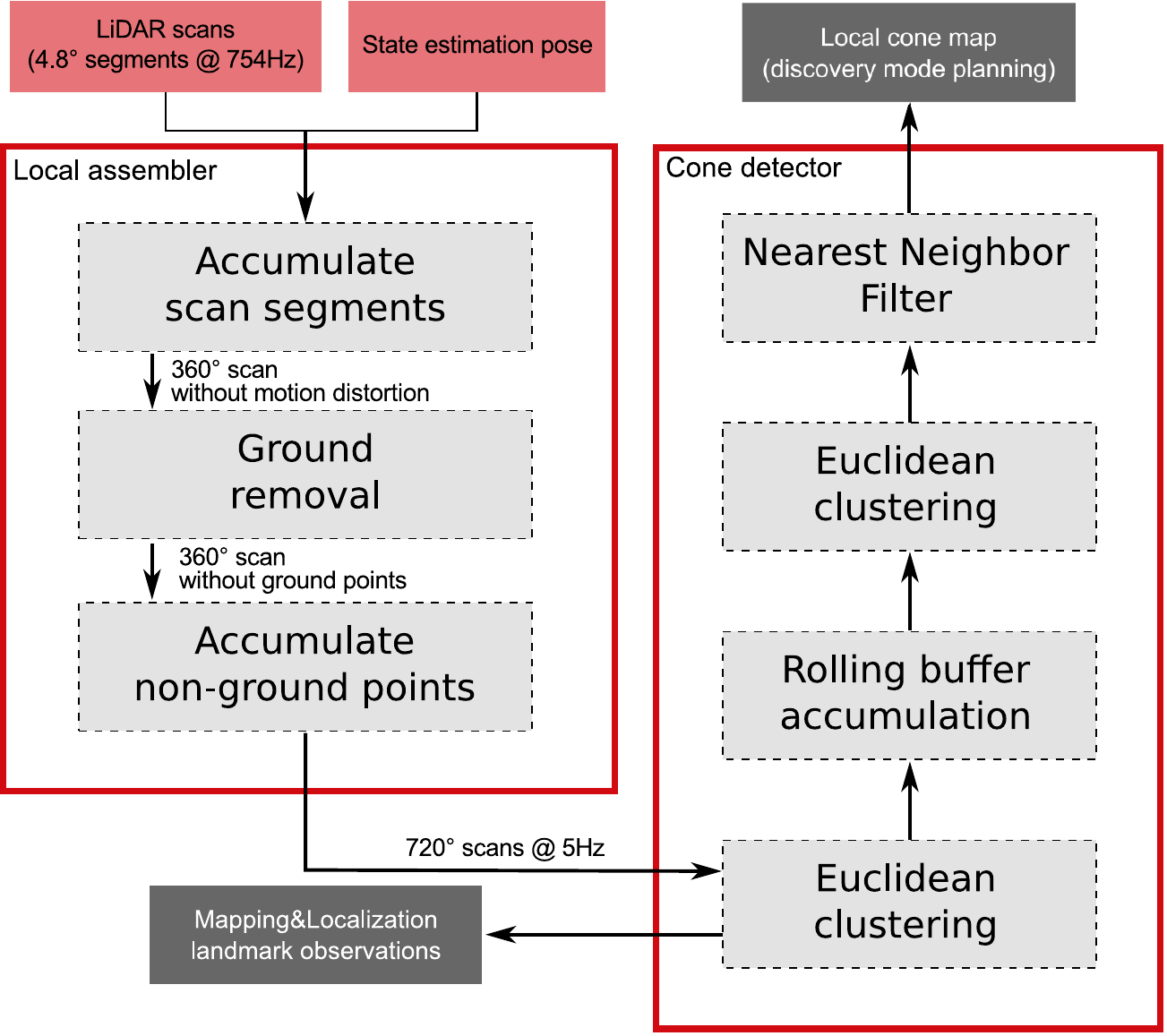}
\end{center}
    \caption{Overview of the LiDAR processing pipeline.}
    \label{fig:lidar_pipeline}
    \vspace{-6mm}
\end{figure}

\begin{figure}
\begin{center}
\includegraphics[width=0.6\columnwidth]{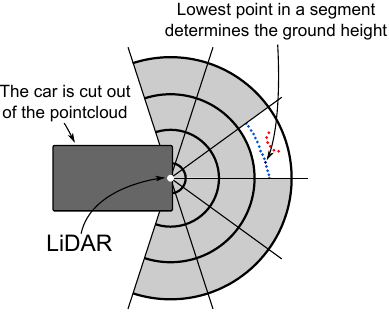}
\end{center}
    \caption{Angular and radial segmentation of the LiDAR scan for ground removal. The number of segments is reduced for illustrational purposes. The experimental setup is run with 12 radial and 7 axial segments.}
    \label{fig:motion-model}
    \vspace{-6mm}
\end{figure}

3D LiDARs are used for detecting cones that mark the race track because of their robustness against variations in illumination. The model used is a Velodyne VLP 16 Puck. Left and right cones are colored blue and yellow respectively. Colors cannot be distinguished from the LiDAR point cloud at an acceptable range, therefore no color information is used. The cone locations are extracted using the pipeline depicted in Figure \ref{fig:lidar_pipeline}. 
The motion distortion is removed from the point cloud by using a velocity estimate provided by the state estimation module. The ground is then removed based on a local flatness assumption. Removal is performed by dividing the scan in segments, as seen in Fig.~\ref{fig:motion-model}, \cite{ground_removal_paper}. Every point that is lower than the lowest point in its segment plus a threshold is removed. Two of these revolutions are accumulated and passed on to the cone detector.

The first step in detecting cones in the ground-free point clouds is Euclidean clustering. The clusters are then classified as cones depending on their size. In an additional filtering step, clusters are rejected using their distance to the LiDAR and the contained points within the cluster.
Cones may not always appear in every scan because of pitching motions and distant cones can fall in between two Velodyne rays. Since multiple LiDAR scans are not fused, this is solved with a second clustering stage. The centroids of the clusters are exported to a new point cloud and the last 10 of these point clouds are stored in a rolling buffer. The combined content of this buffer is processed again with Euclidean clustering. The number of points in the resulting second stage clusters show how often the cone was observed in the last 10 scans. Cones observed twice or more are passed on to the last step.

False positive cones may be detected in areas with uneven terrain or tracks surrounded with high grass. These do not affect the path planning module as they mostly appear outside the track boundaries. However, the computation time of subsequent modules of our system scales with the number of detected cones. A Nearest Neighbour filter is applied to the observed cones to filter out areas with concentration of clusters that are higher than the expected concentration of cones.

\subsection{Mapping \& Localization}
\label{sec:mapping-and-localization}
The maximum range of the perception sensors limits the length of the vehicles path planning horizon. This problem can be overcome by mapping the track and localizing the vehicle within it. As previously mentioned, the track is only marked with cones. The \ac{SLAM} module is designed to accept input from either the LiDAR or camera processing pipeline which ensures safe operation in case of single sensor failure. The track is again assumed to be flat. Only cones are considered as landmarks and other potential features are rejected. There are two distinct phases, corresponding to the previously introduced in Sec. \ref{sec:intro}, \emph{SLAM} and \emph{Localization Mode}. First, the SLAM phase in which the module builds a 2D landmark map of the race track and second, the localization phase where the map is fixed and used to estimate the vehicle pose. The switch from SLAM to localization is performed after a loop closure of the mapped race track is detected. In the following sections, a detailed description of both phases is given.

\subsubsection{SLAM Phase}
The cone observations provided by one of the perception pipelines (camera or LiDAR) are used as landmark inputs. Descriptors cannot be used to aid in data association since the cones are only distinguishable by color (the LiDAR cannot detect the color reliably), geometrically identical and all placed on similar looking asphalt. For this reason, we choose to use FastSLAM \cite{montemerlo2002fastslam}, a Rao-Blackwellized particle filter based SLAM method. Its ability to handle data association on a per particle basis makes it more robust than a SLAM approach that only considers one association hypothesis per time step (e.g., EKF-SLAM). The method also requires a odometry input. The full state estimation pose estimate (see Sec. \ref{sec:state-estimation}) is used while mapping which includes normal GPS. Note that the SLAM pose is not an input to the state estimation, so no information loops are induced.

The map $\mathbf{m}$ is parameterized as a collection of $N$ landmarks. The location of each landmark is estimated using a 2 dimensional EKF ($\boldsymbol{\mu}$, $\boldsymbol{\Sigma}$). Additionally, we record each landmark's observation count $n_{o}$ and missed observation count $n_{m}$ within perception range.
\begin{equation}
\mathbf{m} = \bigl[[ \boldsymbol{\mu}_1, \mathbf{\Sigma}_1, n_{o,1}, n_{m,1}], \ldots, [\boldsymbol{\mu}_N, \mathbf{\Sigma}_N, n_{o,N}, n_{m,N}] \bigr]
\end{equation}

In the particle filter, every particle $\mathbf{Y}^{[i,k]}$ with index $i$ is a combined sample of the vehicle pose $\boldsymbol{\zeta}^{[i,k]}$ and the map $\mathbf{m}^{[i,k]}$ at time $k$.
\begin{equation}
\mathbf{Y}^{[i,k]} = [\boldsymbol{\zeta}^{[i,k]}, \mathbf{m}^{[i,k]}] \quad i=1\ldots P,\, k \in \mathbb{N}
\end{equation}

The particle filter is updated every time a new set of landmark observations $\mathbf{z}$ becomes available. First, the particles poses are propagated using an odometry motion model \cite[pp. 132-139]{probrobotics}, assuming zero mean uncorrelated Gaussian noise on translation and rotation respectively.

Then, observations are associated to existing landmarks in the map. This is done separately for each particle with the maximum likelihood principle. We define a likelihood function $L$ that expresses the likelihood of an observation $\mathbf{z}_i$ coming from a landmark $\mathbf{m}_j$.

\begin{equation}
L(\mathbf{z}_i,\mathbf{m}_j)=\frac{
\exp\left(-\frac{1}{2}({{\mathbf{z}}_i}-{\boldsymbol\mu_j})^T{\boldsymbol\Sigma_j}^{-1}({{\mathbf{z}}_i}-{\boldsymbol\mu_j})
\right)
}{\sqrt{(2\pi)^2|\boldsymbol\Sigma_j|}}
\end{equation}

Observations are assigned to known landmarks in an iterative manner. Mutual exclusion is enforced by using a queue mechanism. If a more likely observation-to-landmark association is found, the previous associated observation is put back into the queue for reconsideration. If an observation cannot be associated with a likelihood of more than the threshold $c$, a new landmark will be initialized for that observation. 

With the now known data association $\mathbf{a}^{[i,k]}$ for every particle, the EKF for each landmark is updated.
Lastly, the weight $w$ of each particle is calculated. The observation likelihoods are incorporated in this weight and the number of new landmarks $l^{[i, k]}$. A penalty $\beta^{[i, k]}$ is added for landmarks that were not observed, but are in the sensor's field of view.
\begin{equation}
w^{[i,k]}=c^{l^{[i,k]}} \cdot \beta^{[i,k]} \cdot \prod_{a_j \in \mathbf{a}^{[i,k]}}L(\mathbf{z}_{a_j}^{[k]}, \mathbf{m}_j^{[i,k]})
\end{equation}
The weights are then normalized.

Resampling of the particle filter is not done at every time step. To determine if resampling is necessary the effective sample size $N_{\mathit{eff}}^{[k]}$ is calculated. Only if this drops below $\frac{3}{4}P$ the particles are resampled using the systematic resampling method \cite{hol2006resampling}.

\begin{equation}
N_{\mathit{eff}}^{[k]} = \frac{1}{\sum_{i=1}^{P}(w^{[i,k]})^2}
\end{equation}

\subsubsection{Loop Closure Detection}
The particle filter based SLAM method has no explicit loop closure detection. To detect the closing of the race track a simple finite state machine method is used. All particles include a loop closure state, which can take three states \emph{Initialized}, \emph{TravelledAway} and \emph{ReturnedHome}. Particles start in the \emph{Initialized} state and, when they move outside a 10m radius from their starting position, are transitioned to the \emph{TravelledAway} state. The \emph{ReturnedHome} state is triggered by coming back within a 5m radius of the starting position, with a heading not deviating more than an angle $\gamma$ from the starting heading. When all particles reached the \emph{ReturnedHome} state and the standard deviation of the pose estimated by all particles drops below 0.1m a closure is assumed. The system then switches to the localization phase.

\subsubsection{Localization Phase}
When the switch is made from mapping to localization the map of the highest weight particle is selected. First the landmarks in this map are pruned. This is done by removing the ones that have an observation ratio $n_o/(n_o+n_m)$ lower than 30\%. Then the track boundaries are determined by linking the landmarks together. The track middle line is isolated from a Voronoi diagram constructed with the track boundaries.

The highest weight particle is now copied to all other particles. The landmark EKF update is disabled, this fixes the map, and the odometry input is switched from full state estimation to an integration of the velocity/wheel sensor and the gyroscope (without GPS). The estimate for the position and heading are extracted from the particle filter by taking the weighted average of all particles. This is fed to the state estimation module for further processing.
\subsection{Safety}
Without a driver, fluela can accelerate from $0-100km/h$ in under $2s$ and corner with up to $1.7g$. This power unfortunately also translates into potential danger. A safety system has thus been devised that guarantees robustness in case of a single mode failure and remains fail-safe in case of combined failures. The system combines safety mechanisms on all levels, from hardware up to the individual software modules.

Starting with hardware, the car has been extended with an Emergency Brake System (EBS) which brakes by default. It can only be released when the HV safety circuit is closed and additionally requires a continuous OK signal from the ECU.
One level higher, the real-time ECU listens for errors on the car's CAN network, monitors the heartbeat of the autonomous master computer and the state of the Remote Emergency System (RES). If the RES is pressed, the car fully engages the brakes within $0.2\si{\second}$, resulting in a deceleration of at least $8\si[per-mode=symbol]{\metre\per\second^2}$ until standstill. With these specifications, a system malfunction at $60\si[per-mode=symbol]{\km\per\hour}$ in a corner would result in the car travelling up to $14\si{\m}$ out of track. This is assuming the safety operator pressed the RES within $0.5\si{\second}$. Eliminating the human reaction time would bring this distance down to $3\si{\m}$.

The autonomous master computer therefore runs a High Level Safety System (HLS), which monitors the heartbeats of each autonomous software module package. The heartbeats carry sequence IDs to detect package loss, time stamps for latency estimation, module load information and a health indicator. The HLS additionally tracks the system resource usage of each module.

On each iteration of the HLS, an anomaly detection algorithm classifies each subsystem as dead or alive. A decision tree then checks if every autonomous function is still covered by at least one package. The car would for example only keep driving in case of a LiDAR pipeline failure if the vision pipeline is still running. The second step is a calculation of the overall system health based on the individual package healths and system resource usage. When driving in \emph{Localization mode}, the top speed is scaled according to the system health. If it falls below a threshold, the car is judged unstable and stopped. If both the decision tree and health threshold deem the car safe to drive, the HLS sends a heartbeat to the ECU and the process repeats.

At the highest level, certain software packages are also allowed to directly trigger the EBS. This would for example happen if the LiDAR detected a large obstacle directly in front of the car whilst racing.
\section{Implementation}\label{sec:implementation}

\subsection{Software setup}
\label{sec:software-setup}
To ensure code quality while keeping the validation process efficient, special attention was paid to the testing methodology and tools.

\subsubsection{Version Control}
Git was used to efficiently code as a team and keep an integral record of the entire project history. 
A three-stage branching model was used to develop and validate functional components. For each new major feature, a Git branch was forked from master into the \textit{Development} stage. Once the new code was written and ready, it was scheduled for testing by moving it to a branch of the \textit{Validation} stage corresponding to the test type and date. After the code had been tested and proven to be stable, it was merged back into the \textit{Stable} stage's only branch: master.

\subsubsection{Continuous Integration}
A Jenkins build server was used to ensure continuous quality control and reveal integration errors early on. Every commit to the aforementioned Git repository was built in a clean workspace to reveal potential errors such as undeclared or clashing dependencies. 

\subsubsection{Simulation}
Gazebo was used in combination with a dynamic model written in python to simulate new features. If new code passed this test, it was ready to be validated on the car. The simulation also proved to be a useful tool for preliminary controller tuning.

\subsubsection{Data Management}
Testing the autonomous system generated a considerable amount of data from different sources. A custom web browser based tool has been developed to efficiently manage all data through one interface.
The information was structured as experiments with report annotation fields and nested test runs. Each test run contains a link to the source code (Git commit hash) that was run and the data (rosbag) it generated.

\subsubsection{Telemetry and diagnostics}
A custom rqt plugin was designed to simplify Telemetry and Diagnostics. When launched in Telemetry mode the GUI would automatically connect the user's laptop to the car, provide a menu to launch and abort autonomous missions, and provide lightweight visuals to illustrate the car state. The GUI's Diagnostics mode allowed hybrid simulation and playback of rosbags. The user could for instance use this to check if a change to path planning had no ill effects on control by replaying all topics excluding the planning and control packages, which would be rerun.

\subsection{Computing assembly}
\label{sec:computing-housing}
A custom housing has been developed to safely embed the computing system in the car. It had to resist light debris and water sprayed up by the left front wheel. Additionally, it had to cool down electronics with a power rating of 170W and shield off EMI from the 480V 3 phase inverters running at 23KHz. These requirements were met with a hermetic, shielded and damped aluminum enclosure. Custom heat spreaders were used to channel the CPU heat from both computers to the walls by conduction. The air inside the box was cooled using 4 wall mounted forced convection heat exchangers. Passive heat sinks mounted on the outside of the box allowed the system to run at full power in steady state when driving at least 5 m/s. At last, the computing housing is mounted to the chassis with shock absorbing thermoplastic elastomer mounts to protect it against vibrations.


\section{Experimental results}\label{sec:results}
\emph{fl\"uela} was tested numerous times to guarantee robustness. It was tested on 8 different locations, on different closed racing tracks ranging from 100 to 500 meters. It was tested under heavy rain for several hours, including FSD, as well as under strong sun over $35^{\circ}C$. Throughout the testing season \emph{fl\"uela} reached over $90km/h$ and $1.7g$ lateral acceleration outperforming amateur drivers in some disciplines.

In this section, we present a performance evaluation based on filed experiments. A video of some experiments can be found at: \url{https://youtu.be/NpLNJ5kC_G0} and a dateset used for some experiments is available at: \url{https://github.com/AMZ-Driverless/fsd-resources#amz_driverless_2017}
\subsection{State Estimation}\label{sec:state-estimation-results}
For the state estimation of the system four parts are evaluated and discussed: accuracy, robustness to outliers, self-diagnosis and delay compensation,.

\subsubsection{Sensor fusion accuracy}
\begin{figure}
\begin{center}
\includegraphics[width=\columnwidth]{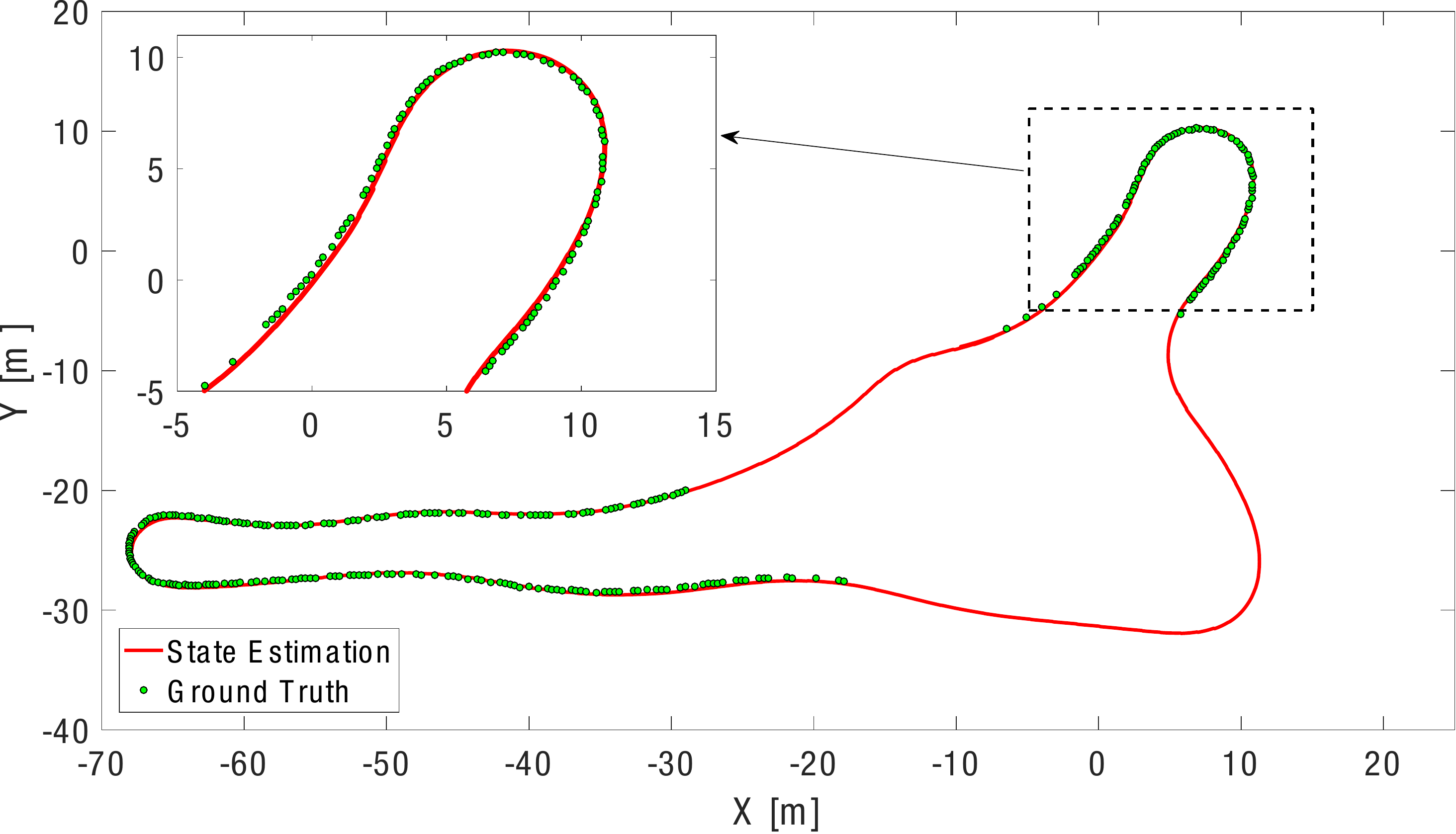}
\end{center}
    \vspace{-5mm}
    \caption{The position from the state estimatior (red) and the ground truth from the laser tracker (green circles) can be seen. The laser tracker could not follow the car at some locations due to its speed. The RMSE is $0.18m$.}
    \label{fig:SensorFusion-leica}
    \vspace{-6mm}
\end{figure}

In order to validate the sensor fusion set up, the position estimate is compared to a ground truth provided by sub-mm precision Leica TotalStation 15i. When compared to ground truth the Root Mean Square Error (RMSE) of the estimated position is $0.18\si{\m}$. See Fig. \ref{fig:SensorFusion-leica}. Note, that the TotalStation sometimes loses the target due to the high speed angular motion required to follow it. These locations are therefore not included in the evaluation.

\subsubsection{Robustness to outliers}
\begin{figure}
\begin{center}
\includegraphics[width=\columnwidth]{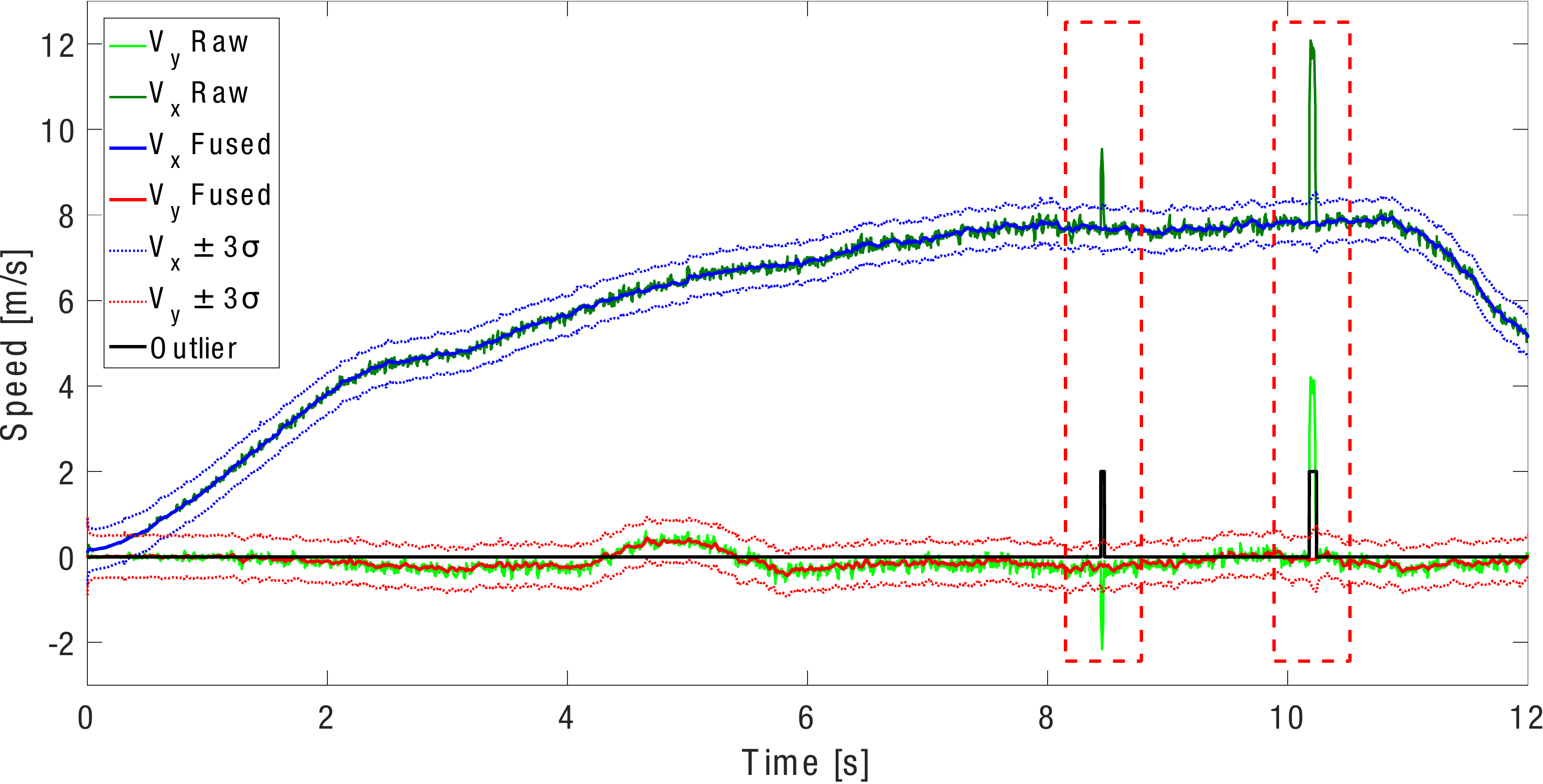}
\end{center}
    \vspace{-5mm}
    \caption{Longitudinal and lateral velocity estimates (blue and red) and their $3- \sigma$ bounds are shown (dotted blue and dotted red). The raw velocity sensor (green) has two faults at $t\approx 8.45 s$ and $t\approx 10.20 s$. The outlier detector (black) spots these faults and rejects them based on the $\chi^2$ test. Since these measurements are not introduced in the EKF, the velocity estimate is not corrupted.}
    \label{fig:ass-outlier}
\end{figure}

\begin{figure}
\begin{center}
\includegraphics[width=\columnwidth]{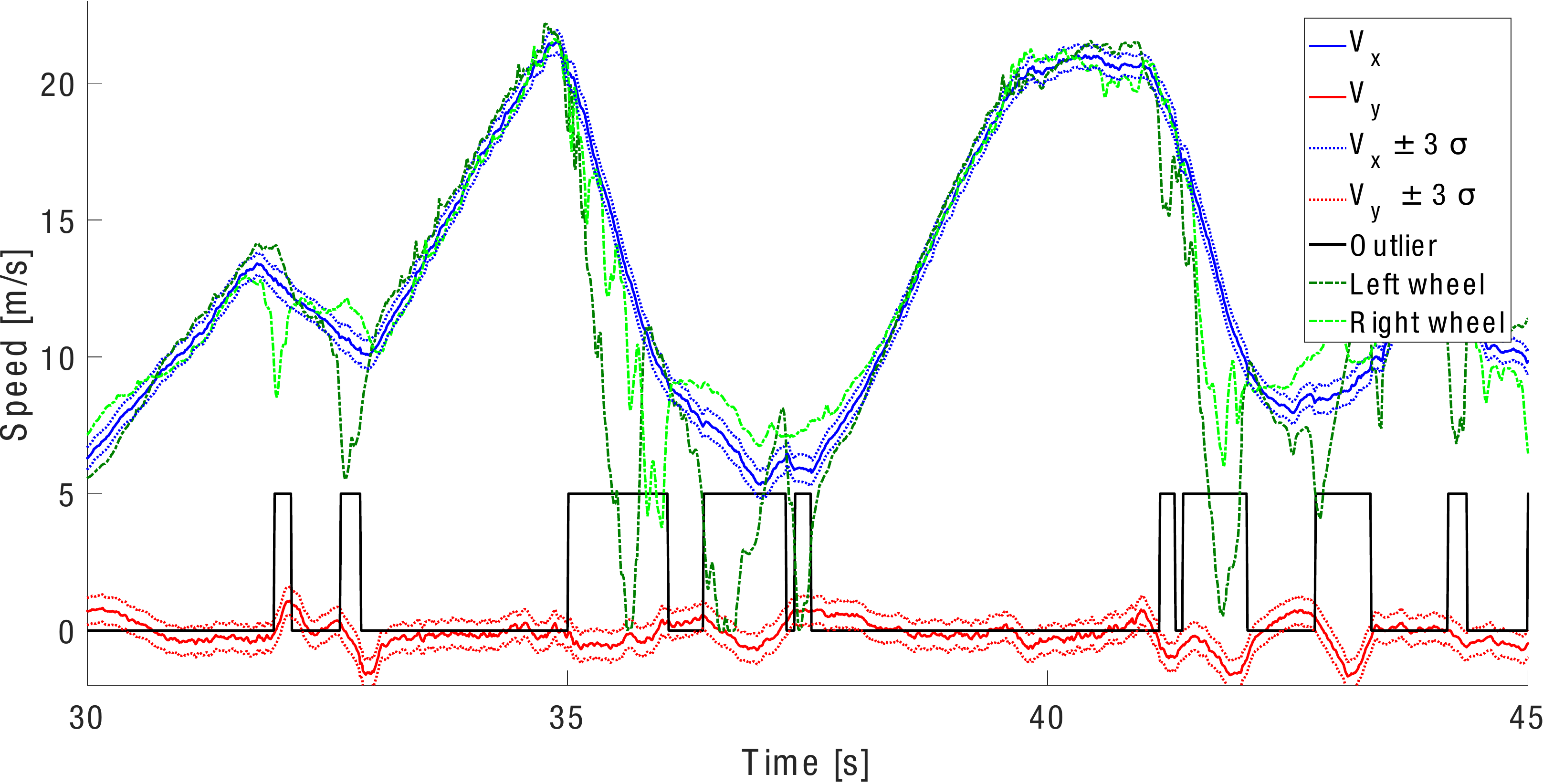}
\end{center}
    \vspace{-5mm}
    \caption{Longitudinal and lateral velocity estimation (blue and red) and their $3- \sigma$ bounds are shown (dotted blue and dotted red). The rear wheels odometriy (green) show that the wheels partially block when braking and turning. The $\chi^2$ test can also be used to detect and reject these measurements (black) that do not represent the speed of the vehicle. }
    \label{fig:odom-outlier}
    \vspace{-6mm}
\end{figure}

The outlier rejection method detailed in Sec.~\ref{sssec:outlier_rejection} was developed to handle the different sensor faults which occurred during the testing phase. The first case of error (Fig. \ref{fig:ass-outlier}) is a velocity sensor which occasionally returns spikes when driving over reflecting surfaces such as water or some road markings. The $\chi^2$ test could reject these cases without exception. The other case presented is the wheel odometry sensor. Due to the high accelerations of the car, one wheel is often blocked when braking and turning at the same time. This can be seen in Fig. \ref{fig:odom-outlier}. The $\chi^2$ test is also used to reject this measurement in these scenarios. It has to be noted that, if wheel odometry is the only velocity source, and if the wheels are constantly blocked due to high accelerations, even with the $\chi^2$ test the velocity estimate deteriorates.
\subsubsection{Self-diagnosis}
\begin{figure}[t]
\begin{center}
\includegraphics[width=\columnwidth]{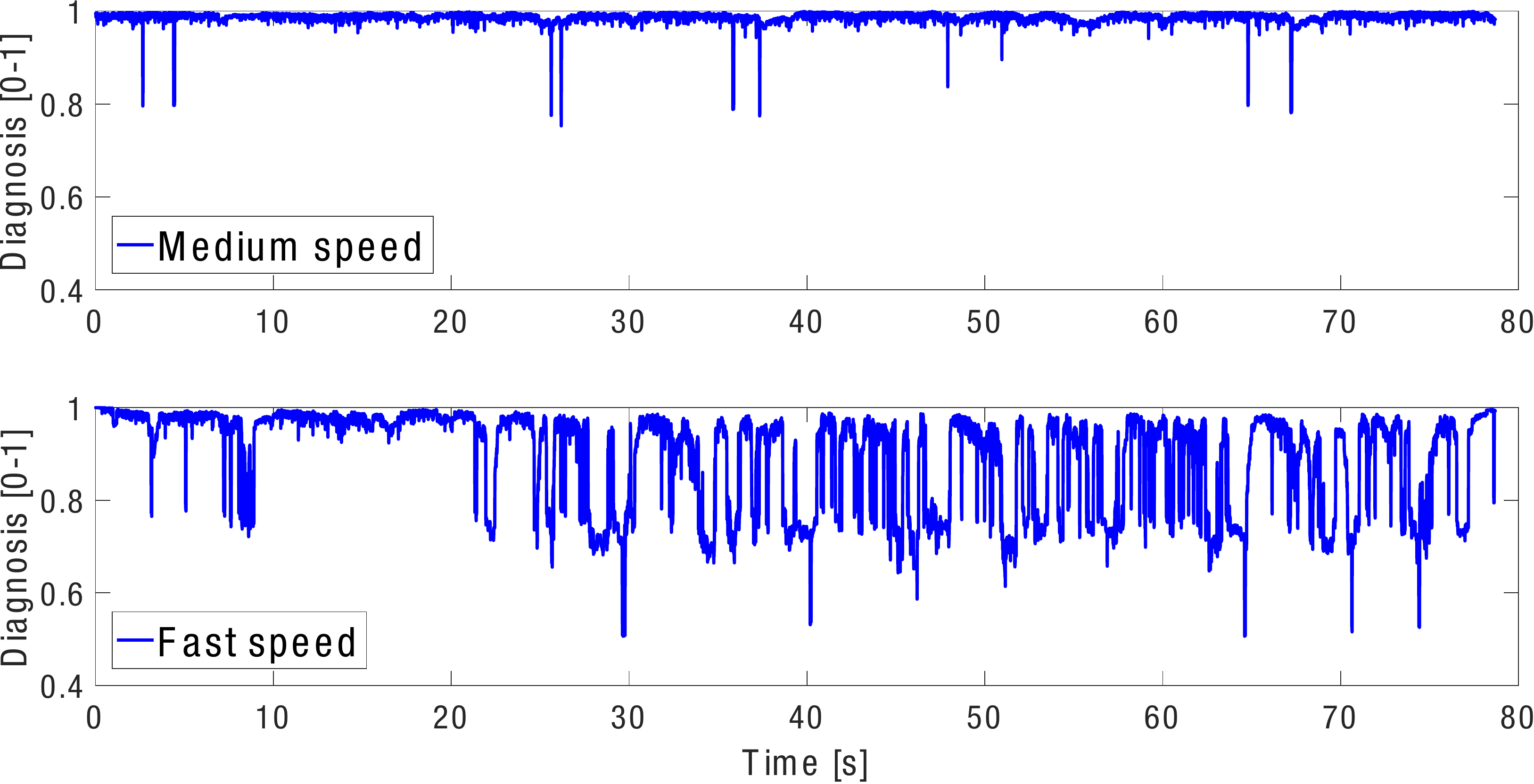}
\end{center}
   \vspace{-5mm}
    \caption{The on-line self-diagnosis is shown for different laps in the same track. The medium speed laps result in a top speed of $30 km/h$ and maximum angular rate of $90^{\circ}/s$. The fast speed laps result in  $80 km/h$ as top speed and $150^{\circ}/s$ as top angular rate. For the medium speed the mean diagnosis is $0.987$ and lowest is $0.75$. For the fast laps the mean is $0.896$ and the lowest $0.5$ }
    \label{fig:health}
\end{figure}

The self-diagnosis results are presented in Fig.~\ref{fig:health} where 1 represents a perfect health and 0 all measurements being outliers. It can be seen that for the same track, the laps reaching $80\si[per-mode=symbol]{\km\per\hour}$ and $150^{\circ}/s$ have a lower health diagnostic. This matches the fact that the medium speed laps have a $0.18\si{\metre}$ RMSE when compared to ground truth and an average health diagnostic of $0.987$ whereas the fast laps have $0.39\si{\metre}$ RMSE and $0.896$ average health diagnostic. This implies that the presented diagnosis method can provide an ad-hock qualitative estimate of the absolute error, without any ground truth information.

\subsubsection{Delay compensation}
\begin{figure}
\begin{subfigure}{}
\includegraphics[width=0.47\columnwidth]{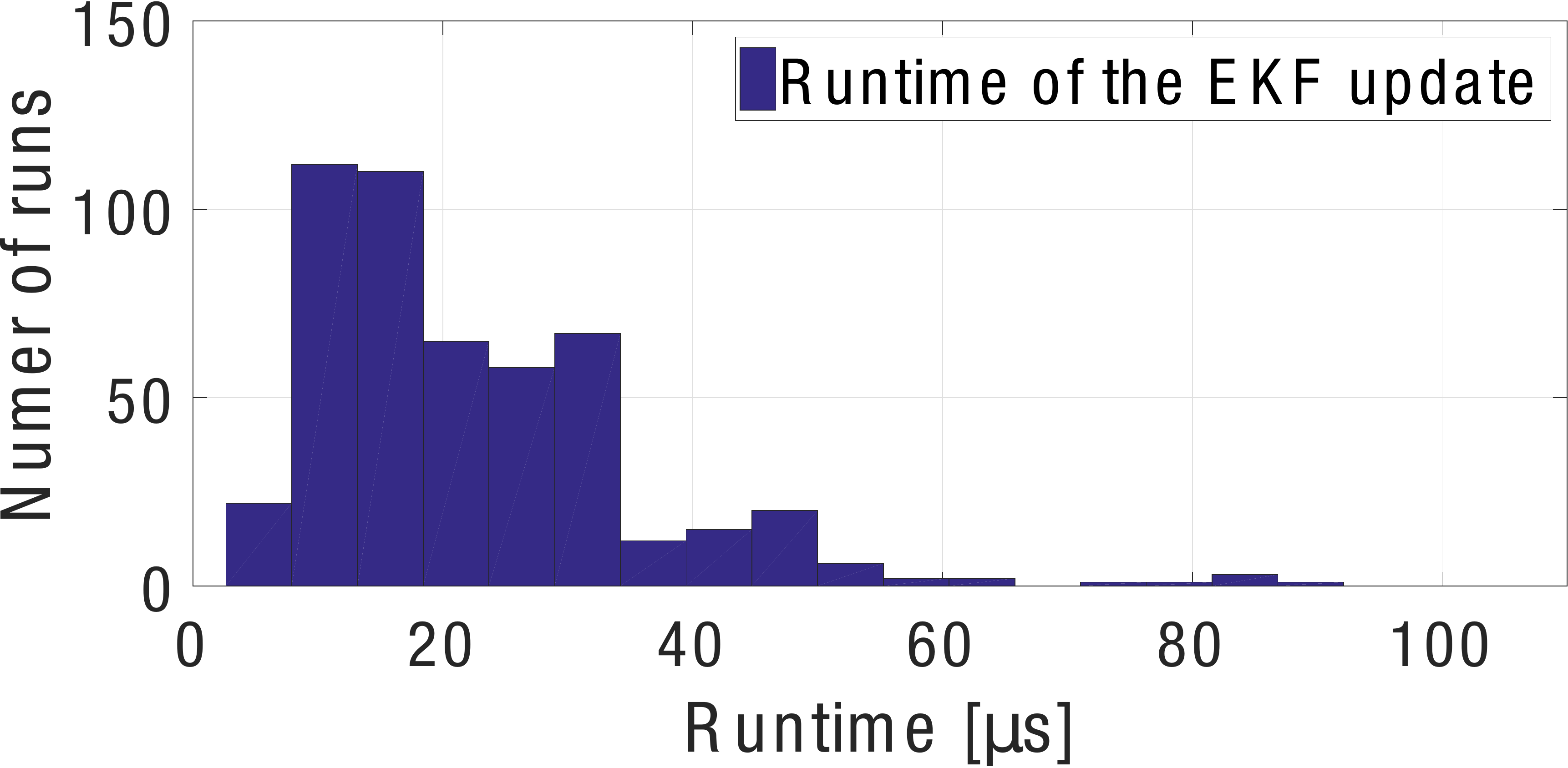} 
\end{subfigure}
\begin{subfigure}{}
\includegraphics[width=0.47\columnwidth]{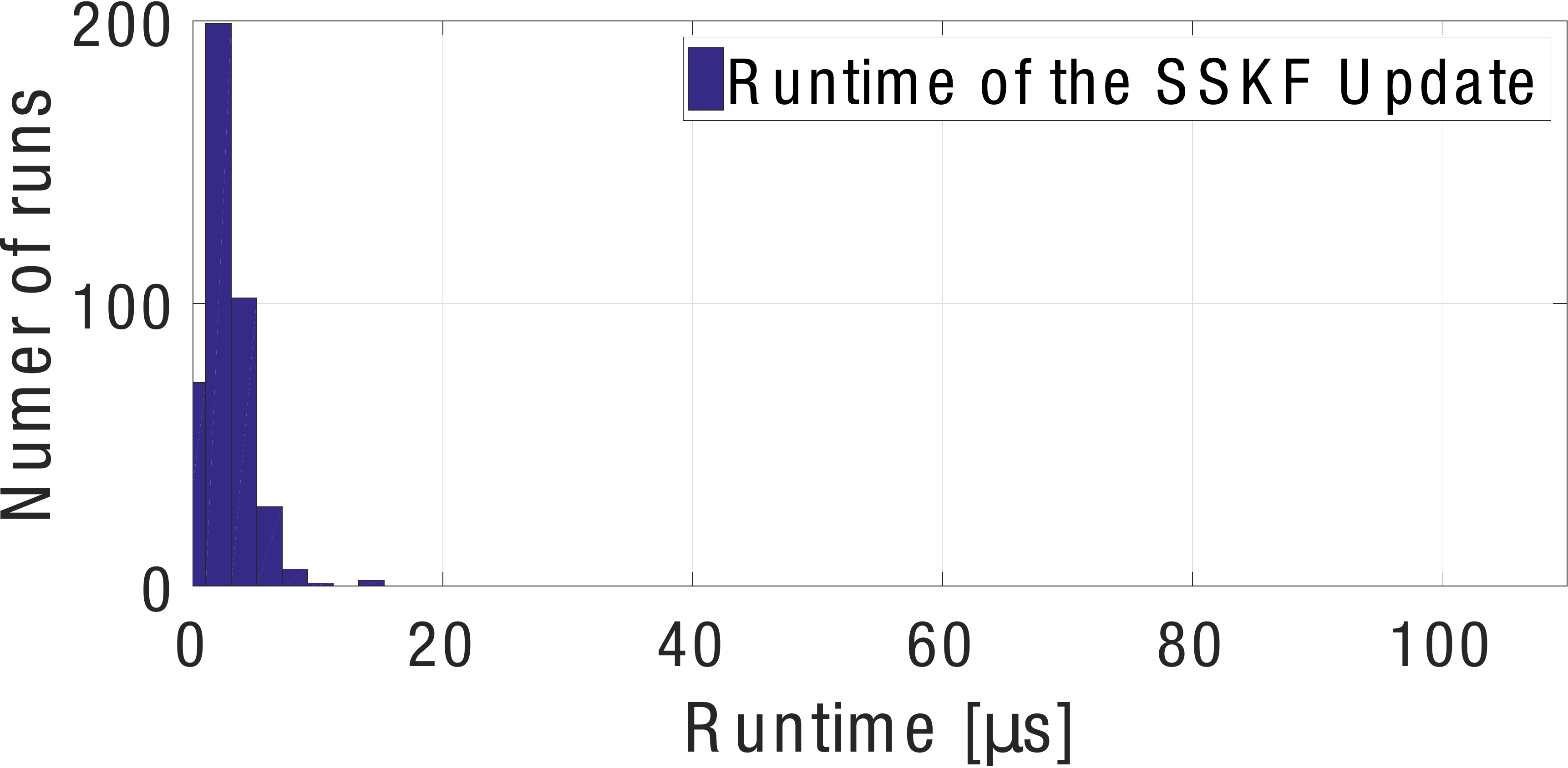}
\end{subfigure}
\vspace{-5mm}
\caption{The runtime of the EKF update (left) and the SSKF update (right) with delays are shown. For this setup with the most delayed measurement varying from $40$ms to $60$ms, the SSKF is around 5 times faster than the EKF.}
\label{fig:EKF_runtime}
\vspace{-6mm}
\end{figure}

In our experiments, the presented approximate delay compensation method is 5 times faster than the EKF intuitive compensation (Fig.~\ref{fig:EKF_runtime}). This varies depending on the most delayed measurement which in our case was from $40ms$ to $60ms$ (4 to 6 EKF iterations). 

\subsection{Mapping and Localization}
\begin{figure}[t]
\begin{center}
\includegraphics[width=\columnwidth]{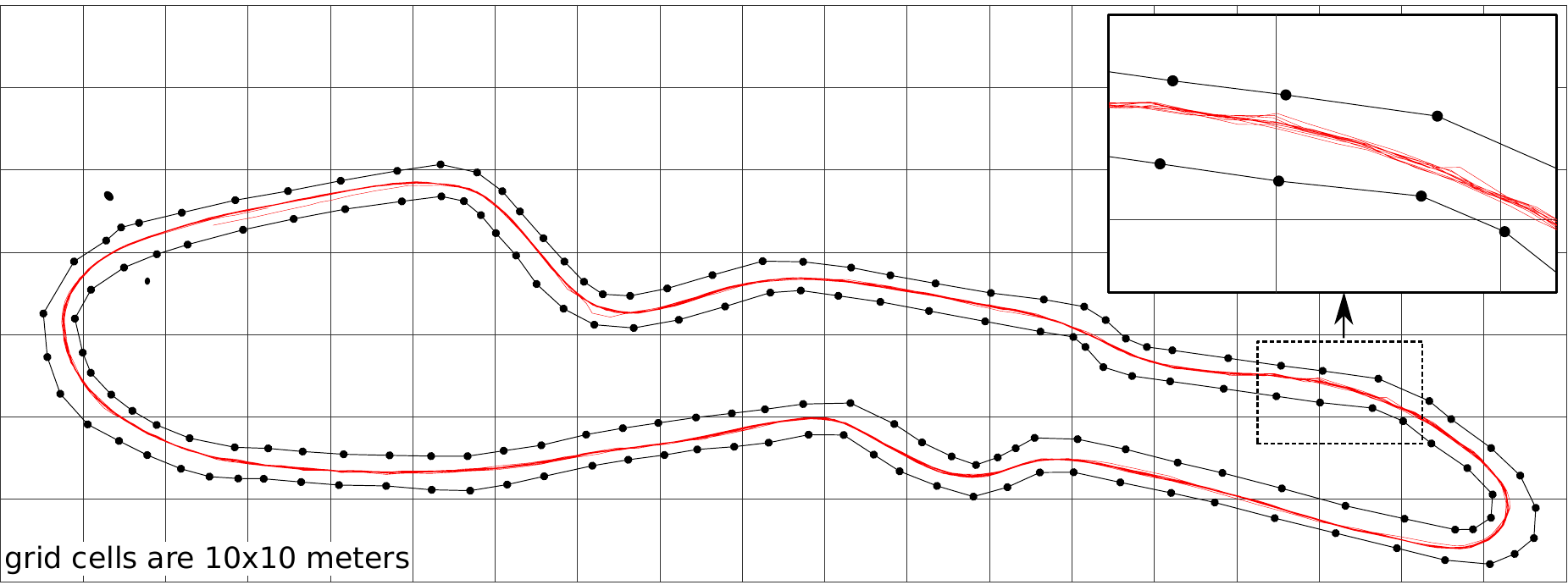}
\caption{A map built using the presented method with 500 particles. The estimated path coming from the mapping and localization module is depicted in red. The 10 laps were driven with a top speed limit of 8 m/s. Black dots are the estimated landmarks, in the top left the timing equipment was detected as a landmark, the rest are all cones. The zoom in shows the sharp edges due to the low update rate (5Hz)}
\label{fig:map}
\end{center}
\vspace{-5mm}
\end{figure}

The map built in real-time during the FSD competition can be seen in Fig.~\ref{fig:map}. The position estimated by the SLAM module is plotted within this map. Linking of the cones to form the boundaries is all done on-board and no manual changes were done to this map other than rotation and scaling for illustration purposes. With this data the particle filter was run at 5Hz using 500 particles with LiDAR cone observations as input. Wheel sensors combined with gyro integration was used as odometry input. The integrated gyro drifted almost $90^\circ$ over the 10 plotted laps, while our localization on the loop-closed map performed robustly throughout. Note, that during the competition it rained heavily, yet our mapping and localization approach performed robustly under these conditions.
    
On an Intel Core i7 7700HQ running at 2.8 GHz the filter update step takes in average 7ms during the mapping phase with a maximum of 29ms. During the localization phase the average computation time is 11ms with a maximum of 28ms. The computation time for the update step of the filter scales linearly with the amount of landmarks. This explains why the average time needed for the mapping phase is lower than for the localization phase.

As the update rate of 5Hz is too low for the control loops, the localization pose is fused with data from the other sensors at a higher rate, an example of this can be seen in Fig.~\ref{fig:SensorFusion-leica}.

\section{Conclusions}\label{sec:conclusion}
This paper presents the state estimation and system integration for an autonomous race car. It is capable of mapping a race track marked with cones using a landmark based SLAM system. Cones are detected with a 3D LiDAR using a two-stage clustering pipeline. The localization output of the SLAM system is used as a virtual position and heading sensor. Together with an INS and velocity sensor, it feeds an EKF based state estimator. Careful vehicle testing revealed the need to extend the state estimator with an outlier rejection and self-diagnosis system. The experiments show that the vehicle can race on unknown race tracks at competitive speeds, even when measurements are distorted due to adverse weather conditions. This perception and state estimation system shows potential to enable future autonomous race car generations to drive at lateral and longitudinal tire limits.

\section*{Acknowledgement}
The authors would like to thank the entire AMZ driverless team for their hard work and passion, as well as all sponsors for their financial and technical support. Furthermore we would like to thank the EU projects TRADR project No. FP7-ICT-609763, and European Union’s Horizon 2020 research and innovation programme under grant agreement No 644227 (Flourish) and 688652 (UP-Drive) and from the Swiss State Secretariat for Education, Research and Innovation (SERI) under contract number 15.0029. and 15.0284 for their support.

\bibliographystyle{ieeetr}

\begin{acronym}
\acro{SLAM}{Simultaneous Localization and Mapping}
\acro{ECU}{Electronic Control Unit}
\end{acronym}

\end{document}